\documentclass[review]{elsarticle}

\usepackage{lineno,hyperref}
\usepackage{mathrsfs}

\usepackage{algpseudocode}  
\usepackage{amsmath}  
\usepackage{amsfonts}

\usepackage{pgfplots}
\usepackage{caption}
\usepackage{subcaption}
\usepackage{microtype}
\usepackage{multirow}
\usepackage{colortbl}
\usepackage{amssymb}
\usepackage[linesnumbered,ruled,boxed,norelsize,vlined,commentsnumbered]{algorithm2e}

\usepackage{amsmath}
\usepackage{amsthm}
\usepackage{adjustbox}
\usepackage{booktabs}

\newcommand{\pgen}{p_{\text{gen}}}
\newcommand{\refine}{p_{\text{refine}}}

\newcommand{\tbl}[1]{\textcolor{black}{#1}}

\journal{}

\biboptions{authoryear}

\begin{document}

\begin{frontmatter}

\title{Image Augmentation Agent for Weakly Supervised Semantic Segmentation}
		
		\author[firstaddress,secondaddress]{Wangyu Wu}\ead{wangyu.wu@liverpool.ac.uk}
            \author[firstaddress,secondaddress]{Xianglin Qiu}\ead{Xianglin.Qiu20@student.xjtlu.edu.cn}
            \author[firstaddress,secondaddress]{Siqi Song}\ead{Siqi.Song22@student.xjtlu.edu.cn}
            \author[thirdaddress]{Zhenhong Chen}\ead{zcheh@microsoft.com}
            \author[secondaddress]{Xiaowei Huang}\ead{xiaowei.huang@liverpool.ac.uk}
		\author[firstaddress]{Fei Ma\corref{mycorrespondingauthor}}\ead{fei.ma@xjtlu.edu.cn}

            \author[firstaddress]{Jimin Xiao\corref{mycorrespondingauthor}}
		\cortext[mycorrespondingauthor]{Corresponding authors} \ead{jimin.xiao@xjtlu.edu.cn}
		
		\address[firstaddress]{Xi'an Jiaotong-Liverpool University, Suzhou, China}
		\address[secondaddress]{University of Liverpool, Liverpool, UK}
            \address[thirdaddress]{Microsoft, Redmond, USA}

\begin{abstract}
Weakly Supervised Semantic Segmentation (WSSS), which utilizes only image-level annotations, has gained considerable attention for its efficiency and reduced cost. However, most existing WSSS methods focus on designing new network structures and loss functions to generate more accurate dense labels, overlooking the limitations imposed by fixed datasets, which can constrain performance improvements. We argue that more diverse trainable images provides WSSS richer information and help model understand more comprehensive semantic pattern. Therefore in this paper, we introduce a novel approach called \textit{Image Augmentation Agent} (IAA) which shows that it is possible to enhance WSSS from data generation perspective. IAA mainly design an augmentation agent that leverages large language models (LLMs) and diffusion models to automatically generate additional images for WSSS. In practice, to address the instability in prompt generation by LLMs, we develop a prompt self-refinement mechanism. It allow LLMs to re-evaluate the rationality of generated prompts to produce more coherent prompts. Additionally, we insert an online filter into diffusion generation process to dynamically ensure the quality and balance of generated images. Experimental results show that our method significantly surpasses state-of-the-art WSSS approaches on the PASCAL VOC 2012 and MS COCO 2014 datasets. Our source code will be released. 
\end{abstract}

\begin{keyword}
Weakly-Supervised Learning\sep Semantic Segmentation\sep Large Language Model \sep Diffusion Model
\end{keyword}

\end{frontmatter}

\section{Introduction}
\label{sec:intro}

\begin{figure*}[t]
\centering
\includegraphics[width=0.8\linewidth]{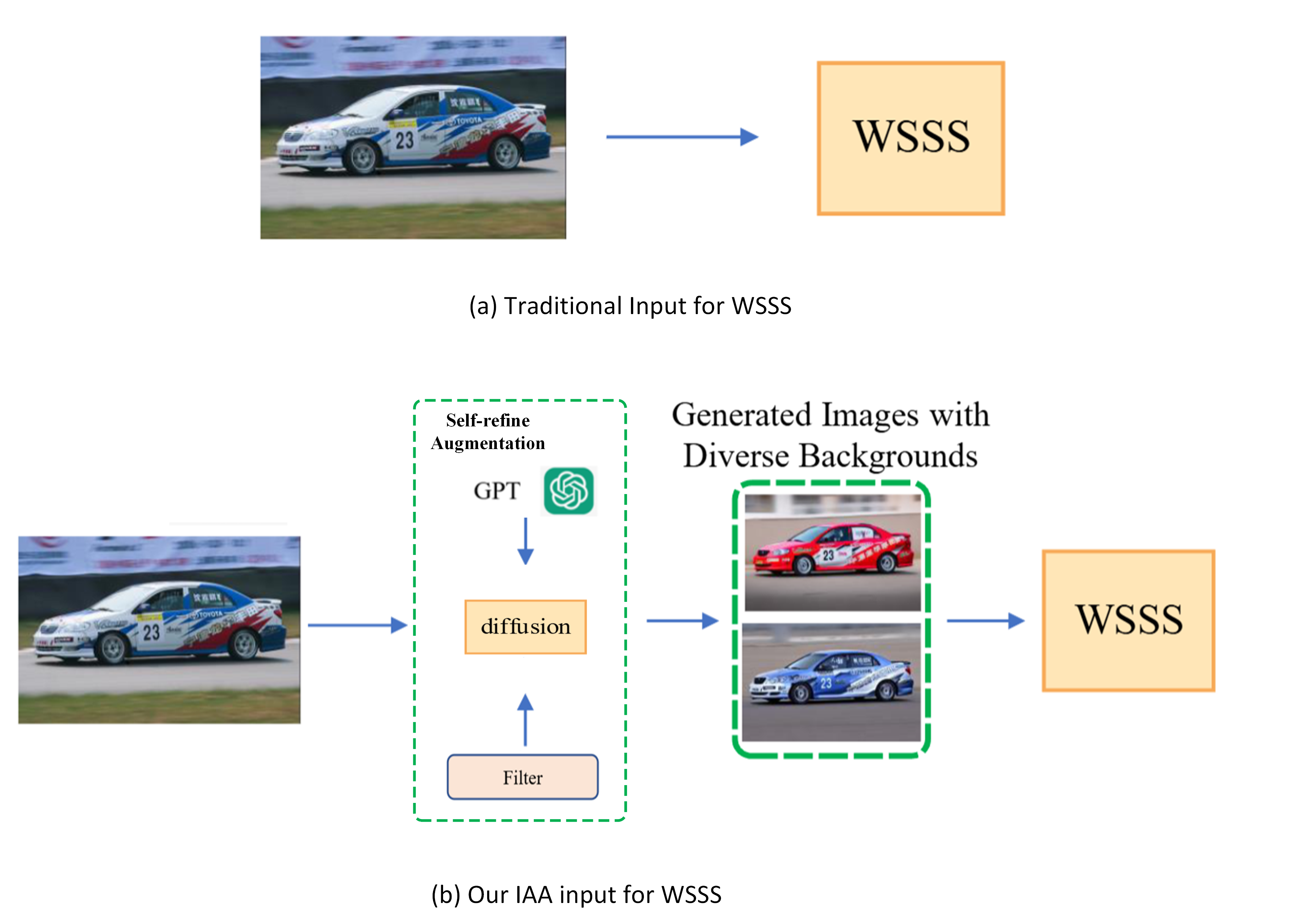}

\caption{\tbl{(a) In the traditional WSSS pipeline, the original image is directly used for training. (b) In our proposed IAA framework, an augmentation agent generates additional synthetic images, which are combined with the original image to enhance training for WSSS.}}
\label{fig:idea}
\vspace{0.4cm} 
\end{figure*}

WSSS leverages image-level labels to perform dense pixel-wise segmentation, making it a cost-effective alternative to fully supervised methods, which often require expensive pixel-wise annotations. The key advantage of WSSS lies in its ability to train segmentation models with minimal supervision, relying solely on image-level labels that provide less granular information but can still guide the model towards accurate segmentation. This approach has gained significant attention in recent years, as it allows for scaling segmentation tasks to large datasets where pixel-wise annotations are unavailable or impractical to obtain. Current mainstream WSSS methods primarily focus on improving the generation of Class Activation Maps (CAMs), which serve as weak supervisory signals for segmentation. These methods typically involve designing novel network architectures and loss functions that enhance the quality and effectiveness of CAMs, which in turn improves segmentation accuracy~\cite{wu5033159generative,ru2023token,zhao2024sfc,yin2023semi,wu2023top}. For example, MCTformer~\cite{xu2022multi} introduced a transformer-based architecture with multi-class tokens to generate class-specific attention maps. This enables a more refined representation of the image, allowing for better localization of objects and more precise CAMs. Similarly, the method presented in~\cite{ru2023token} incorporates a token contrastive loss, which enhances intra-class compactness and inter-class separability. This approach mitigates the problem of over-smoothing in CAM generation, where objects of different classes can be confused with one another due to the lack of fine-grained supervision. \tbl{Despite these advancements, WSSS continues to face two major challenges: (1) the lack of detailed supervision, such as pixel-level annotations, which hampers the model’s ability to accurately localize object boundaries, and (2) the limited scale and diversity of training data, which restricts generalization to unseen categories and complex scenes. These issues result in learning bottlenecks, slow performance gains, and limited robustness.} \tbl{To mitigate these limitations, recent efforts have explored data augmentation using synthetic image generation and external information sources. Notably, IACD~\cite{wu2024image} applied diffusion models to generate additional images for training, as shown in Fig.~\ref{fig:idea}(b). However, IACD suffers from several drawbacks: it employs a fixed background prompt that limits visual diversity, and its post-hoc filtering process may cause an uneven sample distribution, leading to suboptimal dataset quality.}

\tbl{To address the above challenges, we propose a novel \textbf{Image Augmentation Agent (IAA)} that automatically generates diverse, high-quality synthetic training data for WSSS. Our agent integrates LLM-driven prompt generation with a diffusion model and introduces a self-refinement mechanism that enhances both prompt diversity and image quality. The entire pipeline is automated and designed to produce semantically rich and balanced synthetic datasets, as illustrated in Fig.~\ref{fig:idea}(b).}

Our main contributions are summarized as follows:
\begin{itemize}
\item[1)] \tbl{We propose a novel augmentation framework for WSSS that synergizes large language models (LLMs) and diffusion models to generate diverse and semantically consistent training images.}
\item[2)] \tbl{We design a self-refinement mechanism to enhance prompt diversity and improve image quality, addressing the limitations of fixed-prompt and post-filtering-based approaches.}
\item[3)] \tbl{Extensive experiments on PASCAL VOC 2012 and MS COCO 2014 benchmarks show that our method achieves consistent improvements over existing state-of-the-art methods.}
\end{itemize}
\section{Related Work}

\subsection{Weakly-Supervised Semantic Segmentation}
WSSS methods using image-level annotations commonly rely on CAMs as pseudo labels. However, CAMs often highlight only the most discriminative regions of objects, leaving less salient regions unutilized. Various approaches have been proposed to overcome this limitation, including region erasure~\cite{wei2017object}, accumulating attention online~\cite{jiang2019integral}, and mining cross-image semantics~\cite{sun2020mining}. Techniques such as leveraging saliency maps~\cite{lee2021railroad} aim to reduce background interference and discover less obvious object regions. Additionally, contrastive methods~\cite{chen2022self} attempt to activate entire object regions by comparing pixel and prototype representations. Some studies, like~\cite{chang2020weakly,wu2025adaptive,wu2024top,long2025bridging,long2024coinclip,long2023dynamic,ge2025spectral,zhang2024hmt}, enhance WSSS by incorporating more category-specific information or leveraging additional learning signals from the training data. Recent advancements have explored integrating Vision Transformers (ViTs) into WSSS. For instance, MCTformer~\cite{xu2022multi} utilizes ViT attention maps to create localization maps, while AFA~\cite{ru2022learning} leverages multi-head self-attention and affinity modules for propagating pseudo labels. ViT-PCM~\cite{rossetti2022max} pioneers CAM-independent ViT applications for WSSS. \tbl{WeakCLIP~\cite{zhu2025weakclip} introduces a novel CLIP-based text-to-pixel matching framework for WSSS, leveraging vision-language pretraining to refine CAMs and improve segmentation performance.} \tbl{Moreover, recent transformer-based methods have addressed over-smoothing in ViTs to improve segmentation quality. For example, Cheng et al.~\cite{he2023mitigating} propose a calibration framework to mitigate over-smoothing and enhance discriminative power in WSSS, achieving strong performance on benchmark datasets.}

\tbl{These methods primarily optimize network structures or include additional features. In contrast, our work takes a data-centric perspective by focusing on data augmentation through synthetic generation. Rather than modifying model architectures, we aim to enhance WSSS performance by expanding the training corpus in a structured way. Our approach is complementary to existing architectural improvements and can serve as a plug-and-play module compatible with various WSSS pipelines, including transformer-based ones.}

\subsection{Prompt-based Language Models}
Prompt-based learning enhances pre-trained language models (PLMs) by appending task-specific instructions to inputs, allowing the model to better adapt to a wide range of tasks. Early strategies primarily focused on manually crafting prompts that were designed to address specific tasks~\cite{zou2021controllable,zhu2024podb,zhu2025fdnet,li2024high-fidelity,guo2024dual-hybrid}. These manually designed prompts proved effective in certain domains but were inherently limited by their lack of flexibility. As a result, they were difficult to generalize across new or unseen tasks. This limitation spurred research into automating the generation of prompts, allowing for more scalable and adaptable solutions~\cite{shin2020autoprompt,guo2025underwater}. Through the development of automatic prompt generation methods, it became possible to generate prompts dynamically based on the task at hand, enhancing the model's ability to generalize and improving its performance across a variety of domains. One significant advancement in the field was the introduction of continuous prompt optimization~\cite{liu2023gpt}, which further enhanced the adaptability and effectiveness of prompts. By optimizing prompts in a continuous space rather than a discrete one, models can now generate more precise and task-relevant prompts, allowing for greater flexibility in handling a diverse range of tasks. This technique has proven to be effective in improving performance in natural language processing (NLP) tasks, and its principles have since been applied to other areas of machine learning. In addition to their success in text-based applications, PLMs have recently demonstrated significant potential in vision-related tasks. A notable example of this is the use of prompts to enhance few-shot learning for visual recognition~\cite{zhang2023prompt}. By generating task-specific textual prompts that guide the model in interpreting visual input, PLMs have opened new possibilities for cross-domain applications. Unlike traditional methods that rely heavily on manually designed features, this approach leverages the inherent power of language models to understand and generate context-specific instructions that improve task performance. Distinctly, our approach introduces a novel use of PLMs to generate diverse prompts specifically for enhancing WSSS. By generating a variety of prompts that enrich the textual descriptions associated with images, our method aims to improve the performance of WSSS in a way that was not previously explored. To the best of our knowledge, this is the first work to apply self-refinement techniques in PLMs to generate diverse prompts, thereby improving WSSS tasks. This approach not only enriches the textual descriptions used for model training but also contributes to a more effective use of weakly labeled data for segmentation purposes. 

\subsection{Diffusion Models for Data-Efficient Learning}
Diffusion Probabilistic Models (DPMs), introduced by~\cite{sohl2015deep}, have seen substantial progress in image generation. Latent Diffusion Models (LDMs)~\cite{richardson2021encoding} refine this process by performing diffusion in latent spaces~\cite{esser2021taming}, significantly lowering computational requirements. Text-to-image diffusion models, leveraging CLIP~\cite{radford2021learning} and similar pre-trained language models, have achieved remarkable image synthesis results by transforming text into latent representations. Enhanced by methods such as Stable Diffusion~\cite{rombach2022high} and ControlNet~\cite{zhang2023adding}, DPMs now generate high-quality images with precision and minimal artifacts.

Recent studies~\cite{ho2020denoising} demonstrate the utility of DPMs in generating supplementary training data, thereby boosting task performance. \tbl{In parallel to diffusion advancements, recent vision research increasingly explores prompt-driven strategies for improving performance in low-data settings. For instance, M-RRFS~\cite{huang2024m} introduces a memory-based region feature synthesizer for zero-shot object detection, addressing intra- and inter-class variations through prompt-inspired structures. Similarly, UPLVP~\cite{zhang2025unsupervised} presents an unsupervised pre-training framework using language-vision prompts to enhance instance segmentation with limited supervision.}

\tbl{These works highlight the emerging trend of combining prompts and generative models (e.g., diffusion) for data-efficient learning across vision tasks. Motivated by this direction, we propose IAA, which integrates conditional DPMs with GPT-generated prompts to generate high-quality and diverse training images for WSSS. To the best of our knowledge, this represents the first application of conditional diffusion models for WSSS.}

\begin{figure*}[t] 
\begin{center}
   \includegraphics[width=0.8\linewidth]{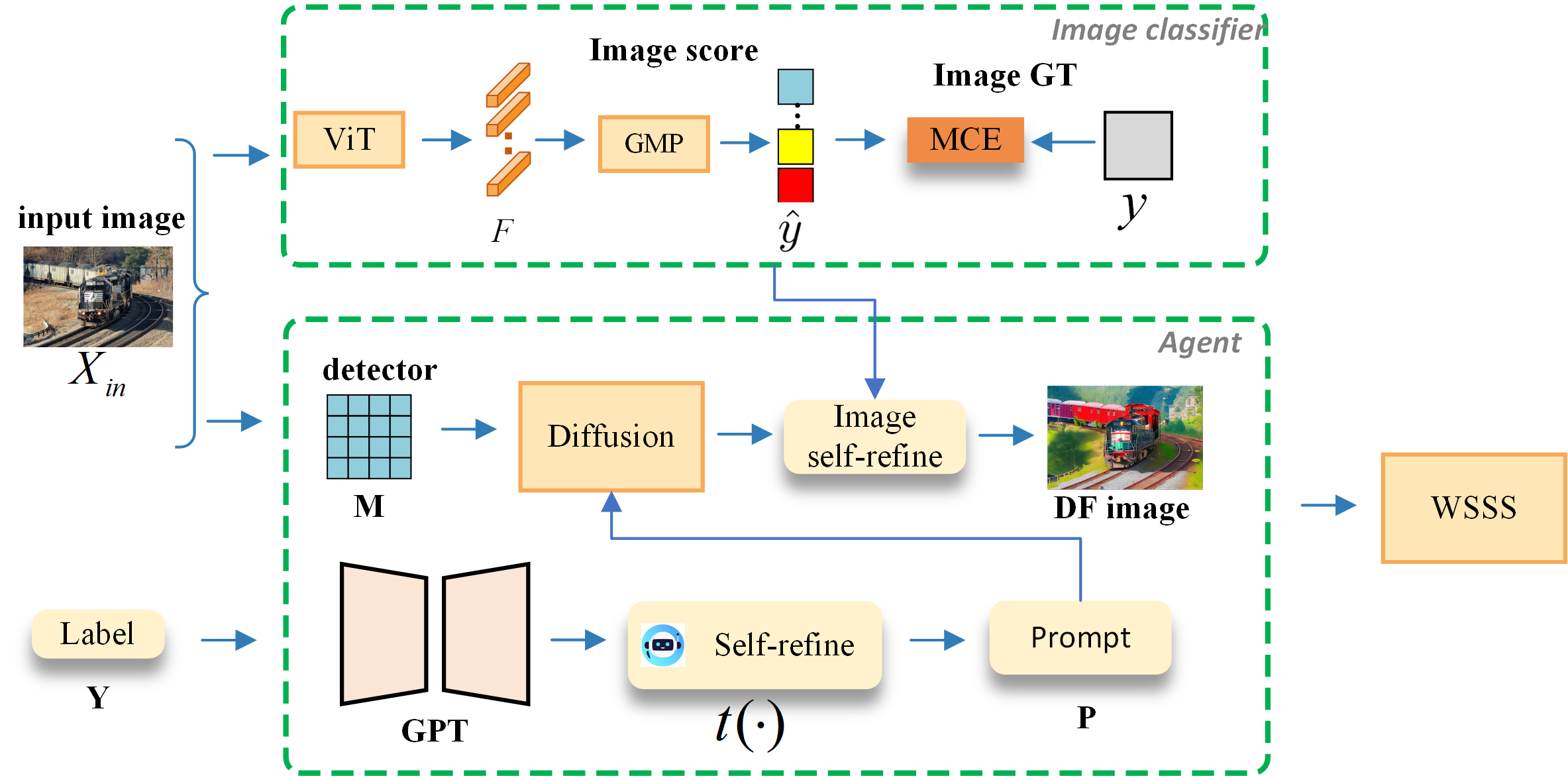}

   \caption{Overall of our IAA. First, IAA uses image-level labels to generate text prompt through self-refine with GPT. Next, the original input image, self-refined GPT prompt and detector map are fed into the diffusion model to generate augmented images. Meanwhile, a pre-trained image classifier serves as a filter, performing image self-refinement to select high-quality images during diffusion generation process. Finally, the selected images are used for WSSS training.}
    \label{fig:IAA}
\end{center}

\end{figure*}
\section{Methodology}
\label{sec:method}

In this section, we will outline the overall architecture and key components of our method. We start with an overview of our IAA in Sec.\ref{sec:Overview}, which integrates multiple agents with ControlNet diffusion and GPT, combined with self-refinement for generating additional images. Subsequently, in Sec.\ref{sec:prompt}, we present our proposed auto-refine Prompt method in the agent to generate diverse background prompts. Finally, in Sec.~\ref{sec:image}, we propose generating augmented data using diffusion with image self-refinement to produce high-quality images. The objective is to generate augmented images through our augmentation agent, increasing the training data size to ultimately enhance semantic segmentation performance.

\subsection{Overall framework} \label{sec:Overview}

As illustrated in Fig.~\ref{fig:IAA}, the components and process of our IAA framework are illustrated. We use ControlNet~\cite{zhang2023adding} as the diffusion~\cite{rombach2022high} backbone and GPT-4o~\cite{hurst2024gpt} as the LLM. We pre-train the image classifier as an image selector using ViT with the original training data. The image class label is input into the LLM module, and we design a self-refinement process to improve prompt quality. The augmentation module uses the training image and the refined prompt from the LLM as input to generate the augmented image. We integrate the classifier as a filter into the diffusion step before generating the image to ensure that each augmented image is of high quality and control the class distribution of generate images. Finally, the training image combined with the augmented image is used as the final input for the WSSS task.

\begin{figure*}[t]
\centering
\includegraphics[width=0.8\linewidth]{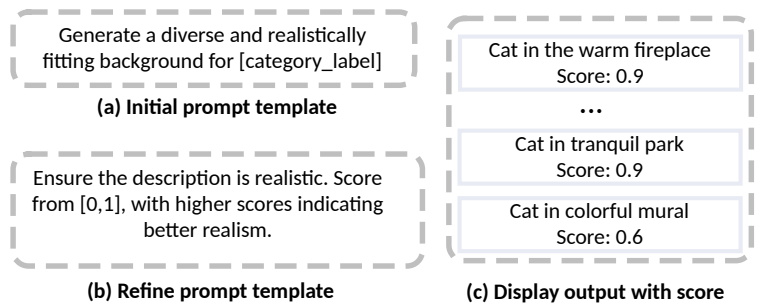}

\caption{(a) the manual template for LLM prompt generation; (b) the prompt template for prompt refinement; (c) the evaluation score of LLM output.}
\label{fig:prompt}
\vspace{0.3cm} 
\end{figure*}
\subsection{Prompt Generation with Self-Refinement} \label{sec:prompt}

\tbl{The motivation behind our prompt generation strategy lies in fully harnessing the extensive world knowledge embedded within LLMs to construct a wide range of background prompts.} \tbl{These prompts serve as critical guides for the diffusion model, enabling it to generate images with diverse styles, semantics, and visual properties.} \tbl{By relying on the generative capabilities of LLMs, we aim to inject rich variability into the prompt space, which in turn supports the creation of more flexible and expressive image outputs. This diversity is especially crucial in vision tasks such as image generation and WSSS, where precise control over visual content is needed.}

\tbl{However, despite their powerful generative capacity, LLMs are not without flaws. A well-known limitation is their inherent instability~\cite{madaan2024self}, which can result in the generation of incoherent, overly generic, or irrelevant prompts that misalign with the intended semantic category. Such low-quality prompts can severely compromise the performance of downstream image generation, particularly when task-specific fidelity is required.}
To address this challenge, we designed a self-refining prompt method that refines the generated prompts iteratively, ensuring that the background prompts are not only reasonable but also closely aligned with the specific category they correspond to. This iterative refinement process allows for the correction of inconsistencies or errors in the initial prompts generated by the LLM, thereby improving their relevance and quality. As shown in Algorithm~\ref{alg:SR}, this self-refining method plays a central role in the prompt generation pipeline. It is represented as the module $t(\cdot)$ in Fig.~\ref{fig:IAA}, where it interacts with the LLM to refine the background prompts continuously. The self-refinement process works by taking the initial prompt generated by the LLM and evaluating its quality based on a predefined set of criteria. If the prompt does not meet the required standards, it is refined further, ensuring that it remains consistent with the category label and maintains its effectiveness in guiding the diffusion model. Through this process, the prompts are iteratively adjusted until they meet the desired level of quality, making them more effective in generating images that are diverse yet relevant to the given task.

This approach not only helps improve the quality of the generated background prompts but also increases the robustness and reliability of the entire image generation process. By ensuring that the prompts align with the current category and are continuously refined, we can guide the diffusion model more effectively and generate images with greater accuracy and diversity. This self-refining method is crucial for handling the complexities of generating high-quality images in tasks such as WSSS, where the ability to generate coherent and diverse images is paramount.

\begin{algorithm}
\caption{{\small{Self-Refine for prompt generation}}}\label{alg:SR}
\SetKwInOut{Input}{Input}
\SetKwInOut{Output}{Output}

\Input{category label $Y$, model \text{GPT-4o}, prompts $\{\pgen,\refine\}$, quality threshold $\epsilon$}
\Output{$P$}
\BlankLine
$y_0 = \text{GPT-4o}~( Initial\_prompt(\pgen, Y))$\;
\Repeat{$score_{y_0} < \epsilon $}{
    $score_{y_0} = \text{GPT-4o}~(Refine\_prompt(\refine, Y,y_0))$\;
    $y_0 = \text{GPT-4o}~( Initial\_prompt(\pgen, Y))$\;
}
$P = y_0$;
\BlankLine
\Return $P$\;
\end{algorithm}
In the WSSS setting, we denote training images as $X_{in}$ and their corresponding labels as $Y$. As depicted in Fig.~\ref{fig:IAA}, for each of the $N$ categories involved in the dataset, a pre-defined template $P_{gen}$ is served as language commands for GPT-4o \cite{hurst2024gpt} to generate intial text output $y_0$, which ensures the relevance and variety of language commands for generating background prompts.

\begin{equation}
    y_{0} = \text{GPT-4o}~(Initial\_prompt(\pgen, Y)),
\end{equation}
Where $p_{gen}$, as shown in Fig.\ref{fig:prompt}(a), is the initial text prompt for generating synthetic samples of category $Y$, and it will be self-refined in our Algo.~\ref{alg:SR}. We use the refined prompt template $\refine$ in Fig.\ref{fig:prompt}b to assess the background prompt quality score. The output $score_{y_0}$, as shown in Fig.\ref{fig:prompt}(c), provides a score for the background. We then select the high-scoring prompt as our final output $P$, which will be used as the prompt in the diffusion models.

 \subsection{Diffusion with Image Self-Refinement} \label{sec:image}

The motivation for using diffusion with image self-refinement is to leverage the ability of controlled diffusion to generate new images that are similar to the original ones while ensuring image quality through our designed image self-refinement mechanism. This combination allows us to maintain the diversity of the synthetic images while ensuring they are of high quality, which is crucial for improving the performance of WSSS. By enriching the training data with additional enhanced images that resemble real-world variations, we aim to improve the model’s ability to handle complex and varied inputs, ultimately improving the final performance of the WSSS task. The primary goal of controlled diffusion is to generate synthetic images that capture the inherent variability found in the original dataset, but with additional alterations introduced by the diffusion process. This approach not only enhances the model’s exposure to a wider variety of visual scenarios but also helps it adapt better to unseen variations, which are typical in real-world applications of WSSS. The image self-refinement step ensures that the generated images maintain high fidelity to the original content, which is crucial for ensuring that the model’s performance is not compromised by the introduction of low-quality synthetic images.

\tbl{To further reduce the domain gap between synthetic and real images, we employ ControlNet to guide the diffusion process using structural conditions such as edge or pose maps derived from the original image. This ensures that the augmented images retain core structural semantics while introducing controlled variations. Although our method does not explicitly use CLIP-based filtering, the use of a structure-aware refinement loop inherently encourages semantic consistency with the original dataset, which helps minimize potential distribution shifts introduced by synthetic augmentation.}

To assess the quality of the generated images, we trained a classifier using the original training data with image-level labels. As shown in Fig.~\ref{fig:IAA}, we use a Vision Transformer (ViT)-based patch-driven classifier to perform the evaluation. The classifier is first trained using the original dataset, which contains images labeled at the image level. The input image $X_{in}$ is divided into $s$ input patches $X_{patch} \in \mathbb{R}^{d\times d \times 3}$ with a fixed size, where $s = \frac{hw}{d^{2}}$ and $h$ and $w$ represent the height and width of the image. The goal is to extract meaningful patch embeddings that capture the local features of the image.

The patch embeddings $F \in \mathbb{R}^{s\times e}$ are then computed using a ViT encoder, which processes the image patches and converts them into high-dimensional representations. A weight matrix $W \in \mathbb{R}^{e\times|\mathcal{C}|}$ is used to map the embeddings into the class space, where $\mathcal{C}$ is the set of categories in the dataset. This weight matrix is applied alongside a softmax function to produce the prediction scores $Z \in \mathbb{R}^{s\times|\mathcal{C}|}$ for each patch:

\begin{equation}
    Z = \text{softmax}(FW),
\end{equation}
where $\mathcal{C}$ is the set of categories in the dataset. The softmax function normalizes the scores, ensuring they represent class probabilities for each patch. Subsequently, global maximum pooling (GMP) is applied to each class to select the highest prediction scores $\hat{y} \in \mathbb{R}^{1\times|\mathcal{C}|}$ among all the patches. This pooling operation helps summarize the class-wise information and produces a single set of image-level prediction scores.

Finally, the vector $\hat{y}$, which contains the image-level prediction scores, is used for image-level classification. The classifier is trained using the multi-label classification prediction error (MCE) loss function. This loss function evaluates the difference between the predicted scores and the ground truth labels for each image. The MCE loss is defined as:

\begin{equation} 
\begin{aligned}\label{eq:MCE}
\mathcal{L}_{MCE} &= \frac{1}{|\mathcal{C}|} \sum_{c \in \mathcal{C}} \text{BCE}(y_c, \hat{y}_c) \\
&= -\frac{1}{|\mathcal{C}|} \sum_{c \in \mathcal{C}} \left[ y_c \log(\hat{y}_c) + (1 - y_c) \log(1 - \hat{y}_c) \right],
\end{aligned}
\end{equation}

where $y_c$ is the ground-truth label for class $c$ and $\hat{y}_c$ is the predicted probability for class $c$. The binary cross-entropy (BCE) loss function is applied for each class independently. Once the classifier has been trained, it can be used to assess the quality of the generated images by evaluating how well they match the ground truth labels. The classifier’s performance in distinguishing between correct and incorrect class predictions serves as an important metric for selecting high-quality generated training data.

Using this classifier, we can filter and select the high-quality augmented images, which will then be incorporated into the final training dataset. These selected high-quality samples are combined with the original dataset to form a robust training set, which can then be used to train the WSSS model. The inclusion of high-quality augmented images significantly contributes to improving the model's ability to perform accurate and precise semantic segmentation, especially in scenarios where labeled data is scarce or expensive to obtain.

Next, we integrate the pre-train classifier into the image self-refinement module. The classifier is incorporated into the diffusion generation step. If the augmented images do not meet the quality criteria, we continue generating images until the desired quality is achieved, rather than applying a filter after all images have been generated. This approach ensures that the augmented images are evenly distributed across all images, preventing the randomness of diffusion from causing some images to have more augmented versions than others. As shown in Fig.~\ref{fig:IAA}, in the diffusion module, we utilize Stable Diffusion with ControlNet~\cite{zhang2023adding} as our generative model. In the data augmentation stage, an input image $X_{in}\in\mathbb{R}^{h\times w\times 3}$, a text prompt $P$ generated by the GPT self-refinement module, and a detection map $M$ are feed into diffusion $\delta(\cdot)$ to generate a new training data $X_{df}$. The detection map is an extra condition (\textit{e.g}., Canny Edge~\cite{ding2001canny} and Openpose~\cite{cao2017realtime}) to control the generation results.

\begin{equation}
    X_{df} = \delta(X_{in}, M, P).
\end{equation}
More details about the data augmentation process are described in Algorithm~\ref{alg:CD}. For images belonging to the 'person' class, we utilize a pose detector map, while for images of other classes, we employ an edge detector map. Subsequently, we utilize GPT-prompt with the detector map to generate augmentation images.

\begin{algorithm}
\caption{{\small{Image Diffusion with Self-Refinement 
}}}\label{alg:CD}
\KwIn{\parbox[t]{8cm}{
    an input image $X_{in}$, an image-level label $Y$}}
\KwOut{a generated image $X_{aug}$}
$P \gets \text{generate\_prompt}(Y)$\\
\For{$t \in \{0, 1, \ldots\}$}{
    \eIf{$``person" \in Y$}{
        $M \gets \text{detect\_map}(X_{in}, \text{human\_pose})$}
    {$M \gets \text{detect\_map}(X_{in}, \text{canny\_edge})$}
    $X_{df} \gets \delta(X_{in}, M, P)$\\
    $score_{df} \gets \text{classifier\_score}(X_{df})$\\
    \If{$score_{df} > \text{high\_quality\_threshold}$}{
        \textbf{break}
    }
}
$X_{aug} \gets X_{df}$\\
\end{algorithm}

\subsection{Final Training Dataset of WSSS} \label{sec:finalinput}

After selecting the high-quality generated training samples, the synthetic dataset $\mathcal{D}_{aug}$ and the original dataset $\mathcal{D}_{origin}$ are combined to form an extended dataset $\mathcal{D}_{final}$ for the training of WSSS. The final training dataset is represented as:

\begin{equation}
\mathcal{D}_{final} = \mathcal{D}_{origin} \cup \mathcal{D}_{aug}.
\end{equation}

This extended dataset $\mathcal{D}_{final}$ serves as a comprehensive training set that not only includes the original data but also incorporates the augmented data generated by our proposed method. The combination of these two datasets is critical for enhancing the diversity and quality of the training data, which in turn improves the performance of the model. 

The synthetic data in $\mathcal{D}_{aug}$ is generated through our IAA method, leveraging advanced augmentation techniques such as LLMs and diffusion models. By generating realistic and varied synthetic images, $\mathcal{D}_{aug}$ complements the original dataset $\mathcal{D}_{origin}$, providing a broader range of visual scenarios for the model to learn from. This is especially valuable for weakly supervised tasks where labeled data is scarce or difficult to obtain. 

The integration of $\mathcal{D}_{aug}$ into the training process allows the model to better generalize to different environments and conditions, improving segmentation accuracy in more challenging scenarios. As a result, the final dataset $\mathcal{D}_{final}$ not only increases the quantity of the training data but also significantly enhances its variety, making it a crucial factor for improving the overall performance of WSSS.

Furthermore, this extended dataset $\mathcal{D}_{final}$ is used to train our segmentation model, allowing it to benefit from both the original data's consistency and the diversity of augmented data. The diversity introduced by $\mathcal{D}_{aug}$ provides the model with new perspectives that are essential for improving segmentation performance, particularly when dealing with complex and varied visual inputs.

\section{Experiments}
\label{sec:Experiments}

In this section, we first present the details of the dataset, evaluation metrics, and implementation. Next, we compare our IAA with state-of-the-art methods on the PASCAL VOC 2012~\cite{everingham2010pascal} and MS COCO 2014 benchmarks~\cite{lin2014microsoft}. Finally, we conduct ablation studies to demonstrate the effectiveness of the proposed method.

\subsection{Experimental Settings}
\begin{table}[t] 
\centering
\caption{\tbl{Semantic Segmentation Performance Comparison (mIoU) on PASCAL VOC 2012.}}
\label{tab:vocseg}
\begin{adjustbox}{width=0.9\linewidth}
\begin{tabular}{@{}llll@{}}
\toprule
Model & Pub. & Backbone & mIoU (\%) \\
\midrule
MCTformer~\cite{xu2022multi} & CVPR22 & DeiT-S & 61.7 \\
SIPE~\cite{chen2022self} & CVPR22 & ResNet50 & 58.6\\
ViT-PCM~\cite{dosovitskiy2020image} & ECCV22 & ViT-B/16 & 69.3 \\
TSCD~\cite{Xu_Wang_Sun_Xu_Meng_Zhang_2023} & AAAI23 & MiT-B1 & 67.3 \\
SAS~\cite{kim2023semantic} & AAAI23 & ViT-B/16 & 69.5 \\
FPR~\cite{chen2023fpr} & ICCV23 & ResNet38 & 70.0 \\
ToCo~\cite{ru2023token} & CVPR23 & ViT-B & 70.2 \\
Zhang \textit{et al.}~\cite{zhang2023weakly} & TIP23&Res2Net-101&71.0\\
WaveCAM~\cite{xu2023wave}& TMM24&ViT-B/16&70.1\\
SFC~\cite{zhao2024sfc} & AAAI24 & ViT-B/16 & 71.2 \\
IACD~\cite{wu2024image} & ICASSP24 & ViT-B/16 & 71.4 \\
PGSD~\cite{hao2024prompt} & TCSVT24 & ViT-B/16 & 68.7 \\
\textbf{IAA} & \textbf{Ours} & ViT-B/16 & \textbf{72.3} \\
\bottomrule
\end{tabular}
\end{adjustbox}
\end{table}

\begin{table}[t] 
\centering
\caption{Semantic Segmentation Performance Comparison (mIoU) on MS COCO 2014.}
\label{tab:cocoseg}
\begin{adjustbox}{width=0.9\linewidth}
\begin{tabular}{@{}llll@{}}
\toprule
Model & Pub. & Backbone & mIoU (\%) \\
\midrule
MCTformer~\cite{xu2022multi} & CVPR22 & Resnet38 & 42.0 \\
ViT-PCM~\cite{dosovitskiy2020image} & ECCV22 & ViT-B/16 & 45.0 \\
SIPE~\cite{chen2022self} & CVPR22 & Resnet38 & 43.6\\
TSCD~\cite{Xu_Wang_Sun_Xu_Meng_Zhang_2023} & AAAI23 & MiT-B1 & 40.1 \\
SAS~\cite{kim2023semantic} & AAAI23 & ViT-B/16 & 44.5 \\
FPR~\cite{chen2023fpr} & ICCV23 & ResNet38 & 43.9 \\
ToCo~\cite{ru2023token} & CVPR23 & ViT-B & 42.3 \\
SFC~\cite{zhao2024sfc} & AAAI24 & ViT-B/16 & 44.6 \\
IACD~\cite{wu2024image} & ICASSP24 & ViT-B/16 & 44.3 \\
PGSD~\cite{hao2024prompt} & TCSVT24 & ViT-B/16 & 43.9 \\
\textbf{IAA} & \textbf{Ours} & ViT-B/16 & \textbf{45.3} \\
\bottomrule
\end{tabular}
\end{adjustbox}
\end{table}

\begin{figure*}[t]
\centering
\includegraphics[width=0.8\linewidth]{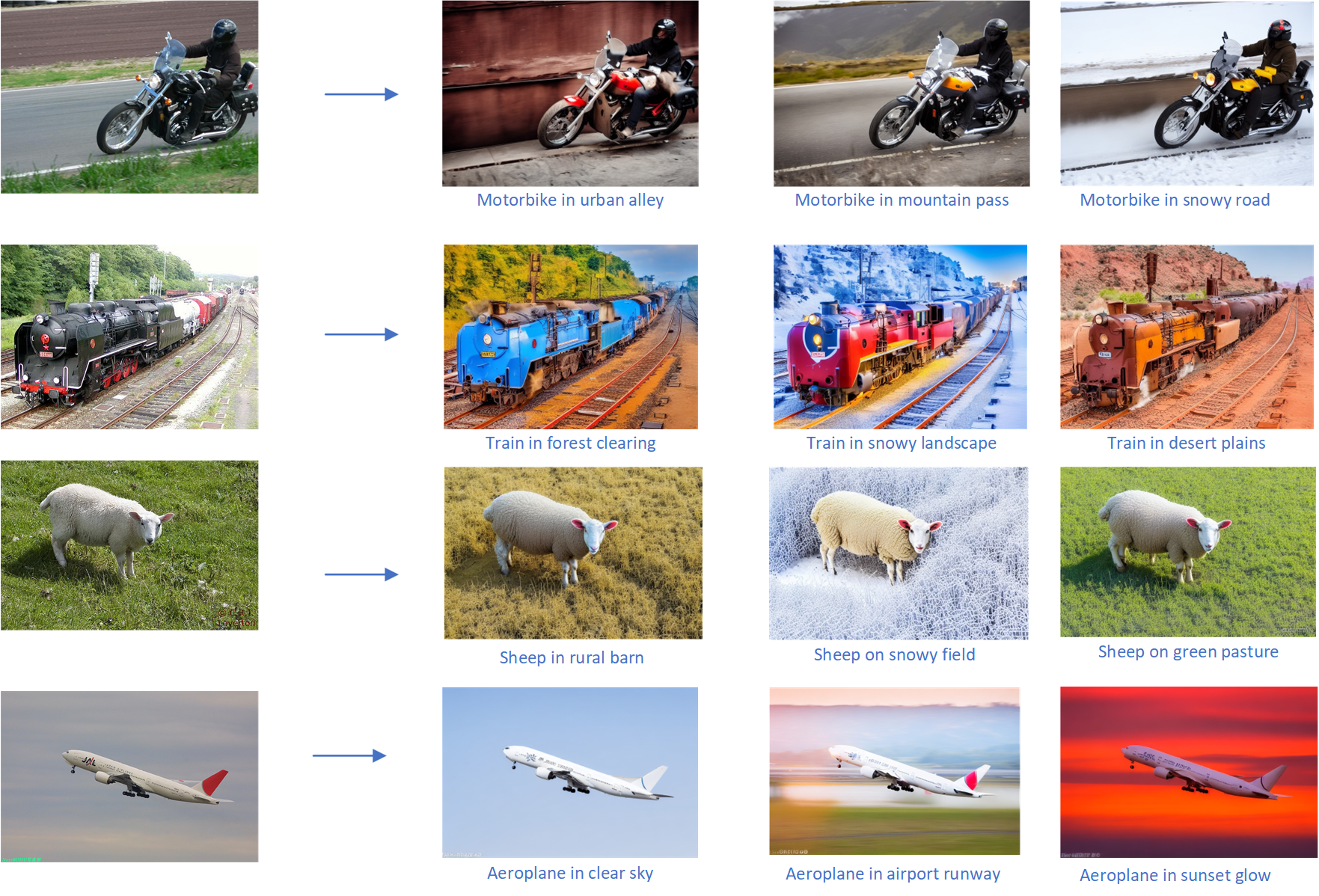}

\caption{Visualizations of ControlNet Generated Images include original training data and augmented images.}
\label{fig:dfimg}

\end{figure*}
\textbf{Dataset and Evaluated Metric.} Our experiments are conducted on two widely used datasets: the PASCAL VOC 2012 dataset~\cite{everingham2010pascal} and the MS COCO 2014 dataset~\cite{lin2014microsoft}. The PASCAL VOC 2012 dataset consists of $21$ categories, including both foreground and background objects, and it is commonly extended using the SBD dataset~\cite{hariharan2011semantic}, which provides additional pixel-wise annotations for semantic segmentation tasks. The MS COCO 2014 dataset, on the other hand, includes $81$ classes, offering a more diverse and complex set of categories for training. For the PASCAL VOC 2012 dataset, we use a total of 10,582 images with image-level annotations for training and a separate validation set of 1,449 images. In the case of the MS COCO 2014 dataset, approximately 82,000 images are utilized for training, and around 40,000 images are reserved for validation, with training images annotated at the image level. 

To evaluate the performance of our methods, we adopt the widely used mean Intersection-over-Union (mIoU) metric. The mIoU provides a comprehensive measure of the model's ability to segment images accurately, considering both the true positives, false positives

\textbf{Implementation Details.} Our IAA method integrates knowledge from pre-trained GPT-4o~\cite{hurst2024gpt}, Stable Diffusion~\cite{rombach2022high}, and ControlNet~\cite{zhang2023adding}. We utilize ViT-B/16 as the ViT model. To facilitate the training of the patch-based image classifier, images are resized to 384×384 as~\cite{kolesnikov2016seed}, and the 24×24 encoded patch features are preserved as input. The model is trained for up to 80 epochs with a batch size of 16, using $\epsilon = 0.9$ as the high-quality threshold for both text and image. Our final training dataset serves as input for the WSSS framework, while keeping all other settings consistent with ViT-PCM~\cite{rossetti2022max}. The experiments were conducted using two NVIDIA 4090 GPUs. Finally, we used the same verification tasks and settings as ViT-PCM~\cite{rossetti2022max}.

\subsection{Final Segmentation Performance}

\begin{figure*}[t]
\centering
\includegraphics[width=0.8\linewidth]{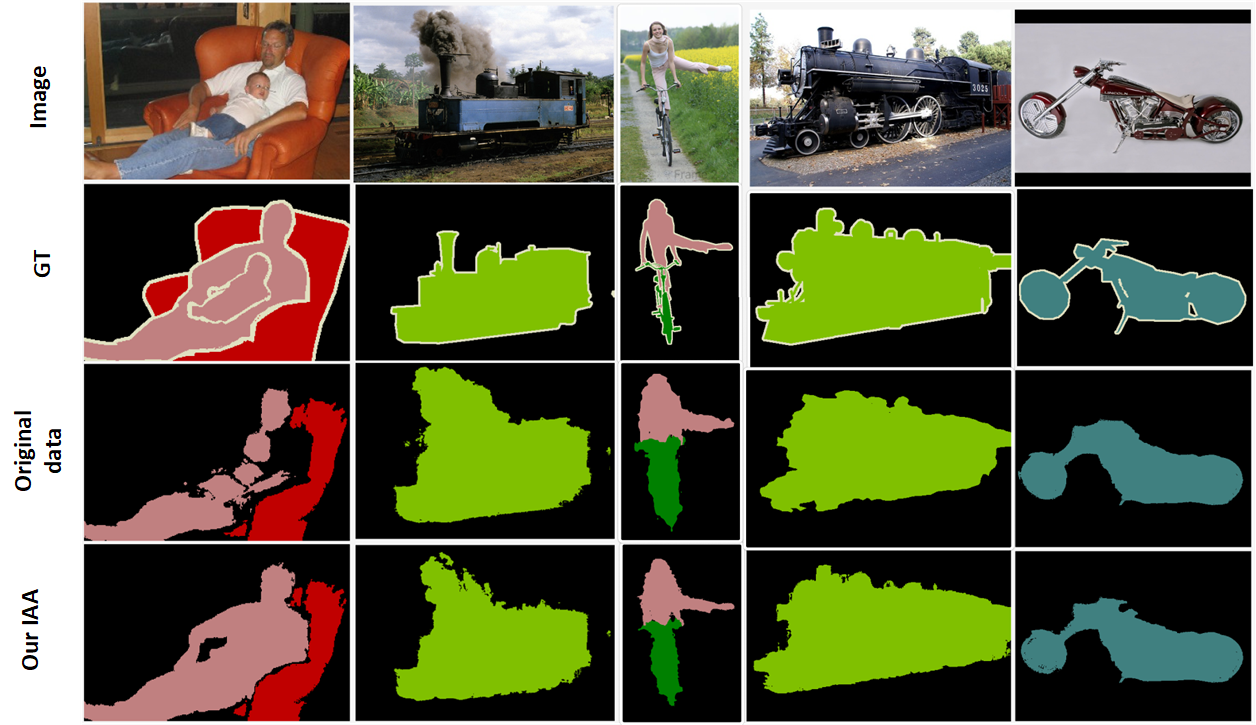}
\caption{Visualization of segmentation results on PASCAL VOC and MS COCO.}
\label{fig:result}
\end{figure*}

\textbf{Comparison with State-of-the-arts.} In this work, we evaluated the final segmentation performance of our proposed IAA method on the PASCAL VOC 2012 and MS COCO 2014 datasets. As presented in Table~\ref{tab:vocseg}, following the approach of other comparative methods, we trained our IAA model using a training set consisting of 10,582 images with only image-level annotations. For validation, we used 1,449 images to assess the performance of our final semantic segmentation results, where our method outperformed existing approaches. For the MS COCO 2014 dataset, we employed a similar strategy, utilizing approximately 82k images for training and around 40k images for validation to evaluate the segmentation performance. As shown in Table~\ref{tab:cocoseg}, our method demonstrated superior results on this dataset as well.

\noindent\textbf{Visualization of results.} Our IAA method effectively leverages an augmentation agent that integrates the power of GPT-4o and diffusion models to enrich the training data for WSSS. By combining the generative capabilities of these advanced models, our approach is able to create diverse and realistic synthetic images that complement the original dataset. This augmentation process not only increases the size of the training data but also enhances its variety, providing the model with a broader range of visual scenarios to learn from, which is crucial for improving segmentation accuracy in complex tasks. Fig.~\ref{fig:dfimg} presents examples of the augmented data that are used as additional training inputs for the model. These synthetic images, generated by the augmentation agent, mirror the characteristics of the original dataset while introducing new variations that would be difficult or costly to obtain through manual annotation. This process is essential in overcoming the limitations posed by the scarcity of labeled data in many real-world applications of WSSS. Furthermore, Fig.~\ref{fig:result} showcases several visualization examples of the segmentation outputs produced by our method. These examples highlight the effectiveness of our approach in generating precise and accurate segmentation maps. It can be observed that our method consistently delivers segmentation results that are more refined and detailed compared to other approaches. The enhanced quality of the segmentation outputs demonstrates the ability of our augmented training data and self-refining modules to significantly improve the model’s performance, particularly in scenarios where weak supervision and limited labeled data are prevalent. Our results provide clear evidence that integrating advanced augmentation techniques with self-refinement mechanisms leads to a substantial improvement in segmentation accuracy and robustness.

\begin{table}[h!]
\centering
\caption{\tbl{Ablation Study on Different Data Augmentation Models: Direct Stable Diffusion vs. Controlled Diffusion with Image Self-Refinement on PASCAL VOC 2012.}}
\begin{adjustbox}{width=\linewidth}
\begin{tabular}{ccccccc}
\toprule
 Original Training Data & Direct Stable Diffusion & Image Self-Refine (Controlled)  &  on Val \\
\midrule
\checkmark &    & & 69.3 \\
\checkmark &  \checkmark  & & 69.0 \\
\checkmark &   & \checkmark\quad\quad\quad\quad\quad & \textbf{72.3} \\

\bottomrule
\end{tabular}
\end{adjustbox}
\vspace{-0.5em}
\label{tab:abs-data}
\end{table}

\subsection{Ablation Studies}

\begin{table}[ht]
\centering
\caption{Ablation studies on main components of the proposed framework on the Pascal VOC 2012 val. 
}
\label{tab:ablation}
\begin{tabular}{cccccc}
\toprule
Original Train & DA & ISR & PSR & mIoU \\
\midrule
\checkmark & & & & 69.3\% \\
\checkmark & \checkmark & & & 69.1\% \\
\checkmark & \checkmark & \checkmark & & 71.7\% \\
\checkmark & \checkmark & \checkmark & \checkmark & 72.3\% \\
\bottomrule
\end{tabular}
\end{table}

\begin{figure}[t]
    \centering
    \includegraphics[width=\linewidth]{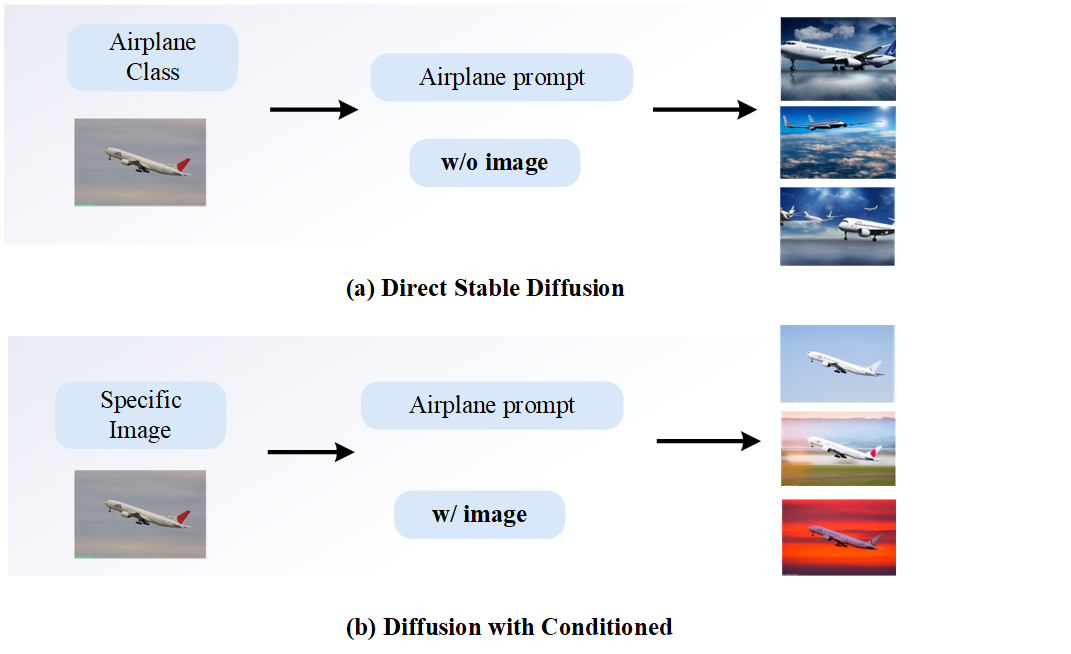 }
    \caption{\tbl{Qualitative comparison of augmented images. Left: Direct stable diffusion result without structural guidance. Right: Image generated using ControlNet with Image Self-Refinement. The controlled method preserves the structure and introduces plausible variations, whereas the uncontrolled result deviates significantly from the original image.}}
    
    \label{fig:ablation_vis}
\end{figure}

We conduct comprehensive ablation studies to evaluate the individual and collective contributions of our proposed components: Prompt Self-Refinement (PSR) and Image Self-Refinement (ISR). As summarized in Tab.~\ref{tab:ablation}, we first integrate Diffusion Augmentation (DA) into the baseline model (original training without augmentation), which results in a marginal performance drop of 0.2\% in mIoU on the validation set. This degradation can be attributed to the inherent randomness of the diffusion process, potentially introducing low-quality or semantically inconsistent augmented images.
To address this limitation, we incorporate Image Self-Refinement (ISR) during the diffusion generation phase, which significantly improves the mIoU by 2.4\%. This substantial gain demonstrates the effectiveness of ISR in enhancing the quality and semantic consistency of the augmented images. Furthermore, the addition of Prompt Self-Refinement (PSR) to diversify the background context of the generated images yields an additional performance boost of 0.6\%, highlighting the importance of diverse and semantically meaningful prompts in guiding the diffusion process.
The progressive performance improvement from 69.3\% to 72.3\% validates the synergistic effect of our proposed components, with ISR playing a particularly crucial role in mitigating the quality issues introduced by the raw diffusion process. These results underscore the significance of both image-level and prompt-level refinement in achieving robust and semantically consistent data augmentation for segmentation tasks. 
\tbl{\paragraph{Detailed Analysis on Image Self-Refinement}In addition, as shown in Table~\ref{tab:abs-data}, we performed an ablation study comparing standard stable diffusion and our \textit{Controlled Self-Refined Diffusion} approach. When using unconditioned stable diffusion, the generated images often differ significantly from the originals, which can introduce noise into the learning process and hinder performance. Even when applying an image selection mechanism, the absence of structural constraints leads to excessive deviation, ultimately reducing segmentation accuracy. In contrast, our \textit{Controlled Self-Refined Diffusion} strategy preserves the core structure of the input image while incorporating subtle variations, resulting in improved model training and better segmentation outcomes. To further illustrate the differences in augmented image quality, we provide a visual comparison in Fig.~\ref{fig:ablation_vis}. The direct diffusion result (left) introduces significant distortion and semantic inconsistency, whereas our Controlled Self-Refined Diffusion result (right) preserves the original structure while introducing meaningful variations. This qualitative difference further supports the effectiveness of our refinement strategy in producing semantically aligned and structurally faithful augmented data.
}

\begin{table}[h]
\centering
\caption{\tbl{(Table 5 in revised version)} \tbl{Normalized CLIP similarity score $(1 + \cos)/2$ between original and augmented images over 100 samples from the \textit{airplane} class on PASCAL VOC.}}
\label{tab:clip_sim}
\begin{adjustbox}{width=0.65\linewidth}
\begin{tabular}{lccc}
\toprule
Image class & Image Size & Augmentation Method & Mean CLIP Similarity \\  
\midrule
Airplane & 100 & Direct Stable Diffusion & 0.750  \\
Airplane & 100 & Controlled Self-Refined Diffusion & \textbf{0.893}  \\
\bottomrule
\end{tabular}
\end{adjustbox}
\vspace{-0.5em}
\end{table}

\tbl{
To quantitatively assess semantic consistency between original and augmented images, we compute CLIP-based cosine similarity scores over 100 randomly sampled \textit{airplane} images. Specifically, we report the normalized similarity score as $(1 + \cos(\theta)) / 2$, where $\cos(\theta)$ is the cosine similarity between the CLIP embeddings of the original and augmented images. As shown in Table~\ref{tab:clip_sim}, images generated using direct stable diffusion exhibit a lower average similarity of 0.750, indicating a larger deviation from the original image distribution. In contrast, images generated via our Controlled Self-Refined Diffusion approach achieve a much higher similarity of 0.893, demonstrating stronger semantic alignment. These results provide quantitative evidence that our strategy effectively reduces the domain gap and produces semantically consistent augmentations.
}
\begin{table}[ht]
\caption{\tbl{\textbf{Effectiveness of Self-Refinement on the \textit{airplane} class.} We report mIoU for three different prompt generation strategies to evaluate the impact of prompt quality and refinement.}}
\centering
\small
\renewcommand{\arraystretch}{1.2}
\begin{tabular}{l|c|c}
\textbf{Prompt Strategy} & \textbf{Class} & \textbf{mIoU} \\
\hline
Fixed Prompt Baseline & airplane & 0.757 \\
LLM Prompt w/o Self-Refine & airplane & 0.763 \\
Ours: Self-Refined Prompt & airplane & \textbf{0.769} \\
\end{tabular}
\label{tab:ablation_airplane}
\end{table}

\tbl{\paragraph{Detailed Analysis on Prompt Self-Refinement}
To assess the necessity of the Self-Refinement module, we conduct an ablation study on the \textit{airplane} class. As shown in Table~\ref{tab:ablation_airplane}, using a fixed prompt yields the lowest performance (0.757 mIoU), indicating that prompt diversity is crucial for effective image generation. When using LLM-generated prompts without refinement, we observe a small performance gain (0.763 mIoU), demonstrating the benefit of leveraging LLM capabilities. }

\tbl{\paragraph{Detailed Analysis on Inference Efficiency}
Although our framework includes additional components during the training phase, it introduces no extra computation overhead during inference. As shown in Table~\ref{tab:inference_time_only}, our method and the baseline both require approximately 0.95 seconds per image, totaling around 23 minutes for the entire PASCAL VOC 2012 validation set of 1449 images. This confirms that our design maintains deployment efficiency, making it suitable for real-world applications despite its training-time complexity.}

\begin{table}[h]
\centering
\caption{\tbl{Comparison of inference time on the PASCAL VOC 2012 val set with and without IAA-enhanced images.}}
\label{tab:inference_time_only}
\begin{adjustbox}{width=0.75\linewidth}
\begin{tabular}{lcc}
\toprule
Setting & Dataset & Inference Time \\
\midrule
w/o IAA image & PASCAL VOC 2012 val & $\sim$23 minutes (0.95 sec/image) \\
w/ IAA image  & PASCAL VOC 2012 val & $\sim$23 minutes (0.95 sec/image) \\
\bottomrule
\end{tabular}
\end{adjustbox}
\vspace{-0.5em}
\end{table}

\tbl{\paragraph{Detailed Analysis on Quality Score Setting}
To validate the choice of $\epsilon = 0.9$, we conducted a human evaluation study across thresholds $\{0.7, 0.8, 0.9, 0.95\}$. For each threshold, we randomly sampled 100 prompts and 100 images from three representative semantic categories (\textit{airplane}, \textit{cat}, and \textit{dog}) in VOC and COCO. Human annotators were asked to assess whether each sample was semantically meaningful and relevant to its class. As shown in Table~\ref{tab:prompt_eval} and Table~\ref{tab:image_eval}, the acceptance rate increases with the threshold and saturates at $\epsilon = 0.9$, where all samples were judged as high-quality. Further increasing $\epsilon$ to 0.95 offers minimal perceptual improvement while significantly increasing generation cost. This confirms that $\epsilon = 0.9$ achieves a good trade-off between quality and efficiency.}

\begin{table}[ht]
\centering
\caption{\tbl{Prompt quality at different $\epsilon$ thresholds (evaluated on 100 randomly sampled prompts per threshold on PASCAL VOC).}}
\label{tab:prompt_eval}
\begin{adjustbox}{width=0.7\linewidth}
\begin{tabular}{lcccc}
\toprule
Category & $\epsilon=0.7$ & $\epsilon=0.8$ & $\epsilon=0.9$ & $\epsilon=0.95$ \\
\midrule
Airplane & 74\% & 88\% & \textbf{100\%} & \textbf{100\%} \\
Cat      & 70\% & 85\% & \textbf{100\%} & \textbf{100\%} \\
Dog      & 76\% & 89\% & \textbf{100\%} & \textbf{100\%} \\
\midrule
Average  & 73.3\% & 87.3\% & \textbf{100\%} & \textbf{100\%} \\
\bottomrule
\end{tabular}
\end{adjustbox}
\end{table}

\begin{table}[ht]
\centering
\caption{\tbl{Image quality at different $\epsilon$ thresholds (evaluated on 100 randomly sampled images per threshold on PASCAL VOC).}}
\label{tab:image_eval}
\begin{adjustbox}{width=0.7\linewidth}
\begin{tabular}{lcccc}
\toprule
Category & $\epsilon=0.7$ & $\epsilon=0.8$ & $\epsilon=0.9$ & $\epsilon=0.95$ \\
\midrule
Airplane & 68\% & 89\% & \textbf{100\%} & \textbf{100\%} \\
Cat      & 71\% & 86\% & \textbf{100\%} & \textbf{100\%} \\
Dog      & 69\% & 90\% & \textbf{100\%} & \textbf{100\%} \\
\midrule
Average  & 69.3\% & 88.3\% & \textbf{100\%} & \textbf{100\%} \\
\bottomrule
\end{tabular}
\end{adjustbox}
\end{table}

\section{Conclusion}

In this work, we propose the IAA method for WSSS, addressing key challenges through innovative data augmentation. Unlike traditional methods relying solely on original training data, our approach introduces an augmentation agent that leverages diffusion models and LLMs to generate diverse synthetic images consistent with the original dataset. These images enrich the training data, providing greater variety and enabling the model to learn from a broader range of visual scenarios. By seamlessly integrating augmented images with the original data, our method significantly expands the dataset without additional manual labeling, mitigating the issue of limited labeled data. This expansion not only improves model performance but also reduces overfitting by exposing the model to more diverse and realistic contexts. Additionally, we design two key modules to enhance image quality and generalization: the self-refinement prompt and the self-refinement image module. The self-refinement prompt iteratively improves generated background prompts, ensuring task relevance, while the self-refinement image module iteratively optimizes generated images to meet desired characteristics. Together, these modules enhance the data augmentation process, producing higher-quality and more reliable training samples for improved segmentation performance.

Ultimately, our data augmentation method demonstrates SOTA results in weakly supervised semantic segmentation, setting a new benchmark for the field. By combining cutting-edge techniques in data generation, prompt refinement, and image quality enhancement, we offer a robust solution for overcoming the challenges of limited labeled data, improving model performance, and achieving more accurate segmentation results. This approach opens up new avenues for future research in WSSS, particularly in expanding the capabilities of weakly supervised models to handle a wider variety of real-world scenarios.

\tbl{In future work, we plan to extend our evaluation to additional datasets beyond PASCAL VOC and COCO, especially those involving more diverse, complex, and real-world scenes. This will help assess the robustness and generalization capacity of our method. Furthermore, we aim to explore the adaptation of our augmentation framework to downstream tasks such as object detection and instance segmentation, and to investigate its effectiveness in low-resource or domain-shifted environments. These directions will provide deeper insights into the scalability and applicability of our approach in broader practical contexts.}

\section{Acknowledgements}
This work was supported by the National Natural Science Foundation of China (No. 62471405, 62331003, 62301451), Suzhou Basic Research Program (SYG202316) and XJTLU REF-22-01-010, XJTLU AI University Research Centre, Jiangsu Province Engineering Research Centre of Data Science and Cognitive Computation at XJTLU and SIP AI innovation platform (YZCXPT2022103).
\bibliography{mybib}

\end{document}